\documentclass[letterpaper, 10 pt, conference]{ieeeconf}  

\IEEEoverridecommandlockouts                              

\usepackage{cite}
\usepackage{amsmath,amssymb,amsfonts}
\usepackage{algorithmic}

\usepackage{graphicx}
\usepackage{adjustbox}
\usepackage{subcaption}
\usepackage{caption}

\usepackage[dvipsnames]{xcolor}
\usepackage{tikz}
\usepackage{dirtree}

\usepackage{academicons}
\definecolor{orcidlogocol}{HTML}{A6CE39}

\usepackage{tabularx}
\newcolumntype{x}[1]{!{\centering\arraybackslash\vrule width #1}}
\newcolumntype{L}[1]{>{\raggedright\arraybackslash}p{#1}} 

\usepackage{adjustbox}
\usepackage{array}
\usepackage{multirow}

\newcolumntype{R}[2]{%
	>{\adjustbox{angle=#1,lap=\width-(#2)}\bgroup}%
	l%
	<{\egroup}%
}

\usepackage{arydshln}

\usepackage{booktabs}

\usepackage[inkscapelatex=false]{svg}

\definecolor{lime}{HTML}{A6CE39}

\DeclareRobustCommand{\orcidicon}{%
	\begin{tikzpicture}
		\draw[lime, fill=lime] (0,0) 
		circle [radius=0.16] 
		node[white] {{\fontfamily{qag}\selectfont \tiny ID}};
		\draw[white, fill=white] (-0.0625,0.095) 
		circle [radius=0.007];
	\end{tikzpicture}
	\hspace{-2mm}
}

\newcommand{\orcidWalter}{\href{https://orcid.org/0000-0003-4565-1272}{\orcidicon}}
\newcommand{\orcidChristian}{\href{https://orcid.org/0000-0003-4822-2844}{\orcidicon}}
\newcommand{\orcidTung}{\href{https://orcid.org/0000-0002-7908-6112}{\orcidicon}}
\newcommand{\orcidKnoll
}{\href{https://orcid.org/0000-0003-4840-076X}{\orcidicon}}
\usepackage{textcomp}

\usepackage{siunitx}
\def\BibTeX{{\rm B\kern-.05em{\sc i\kern-.025em b}\kern-.08em
		T\kern-.1667em\lower.7ex\hbox{E}\kern-.125emX}}

\makeatletter 
\newcommand{\linebreakand}{%
\end{@IEEEauthorhalign}
\hfill\mbox{}\par
\mbox{}\hfill\begin{@IEEEauthorhalign}
}    

\makeatletter
\newcommand*{\emails}[2][@tum.de]{%
	\def\@tempa{\@gobble}%
	\@for\qrr@email:=#2\do{%
		\edef\@tempb{\noexpand\href{mailto:\qrr@email #1}{\qrr@email}}%
		\edef\@tempa{\unexpanded\expandafter{\@tempa}{, }\unexpanded\expandafter{\@tempb}}}%
	\{\@tempa\}#1%
}    

\usepackage{capt-of}
\usepackage{etoolbox}
\usepackage{wasysym}

\usepackage{hyperref} 
\hypersetup{
	bookmarks=false,
	colorlinks=true,
	citecolor=green,
	linkcolor=blue,
	filecolor=magenta,      
	hypertexnames=true,
	urlcolor=cyan,
	pagebackref=true,
	pdftitle={A9-Intersection-Dataset},
	pdfpagemode=FullScreen
}
\usepackage[capitalize]{cleveref}

\crefname{figure}{Fig.}{Figs.}
\Crefname{figure}{Figure}{Figures}
\crefname{section}{Sec.}{Secs.}
\Crefname{section}{Section}{Sections}
\crefname{table}{Tab.}{Tabs.}
\Crefname{table}{Table}{Tables}
\crefname{equation}{Eq.}{Eqs.}
\Crefname{equation}{Equation}{Equations}

\title{\LARGE \bf
	A9 Intersection Dataset: \\All You Need for Urban 3D Camera-LiDAR Roadside Perception
}

\author{
	Walter Zimmer*\orcidWalter \and Christian Creß*\orcidChristian \and Huu Tung Nguyen*\orcidTung \and Alois C. Knoll\orcidKnoll
	\thanks{The authors are with the Chair of Robotics, Artificial Intelligence and Real-time Systems, TUM School of Computation, Information and Technology (CIT), Technical University of Munich, Munich, Germany. Contact: 
		{\tt\small walter.zimmer@tum.de, christian.cress@tum.de}
	}
	\thanks{*These authors contributed equally.}
}

\begin{document}
	
	\makeatletter
	\let\@oldmaketitle\@maketitle
	\renewcommand{\@maketitle}{\@oldmaketitle
		\includegraphics[height=8.0cm,width=1.0\linewidth,trim={0cm 2.5cm 0cm 0cm}]{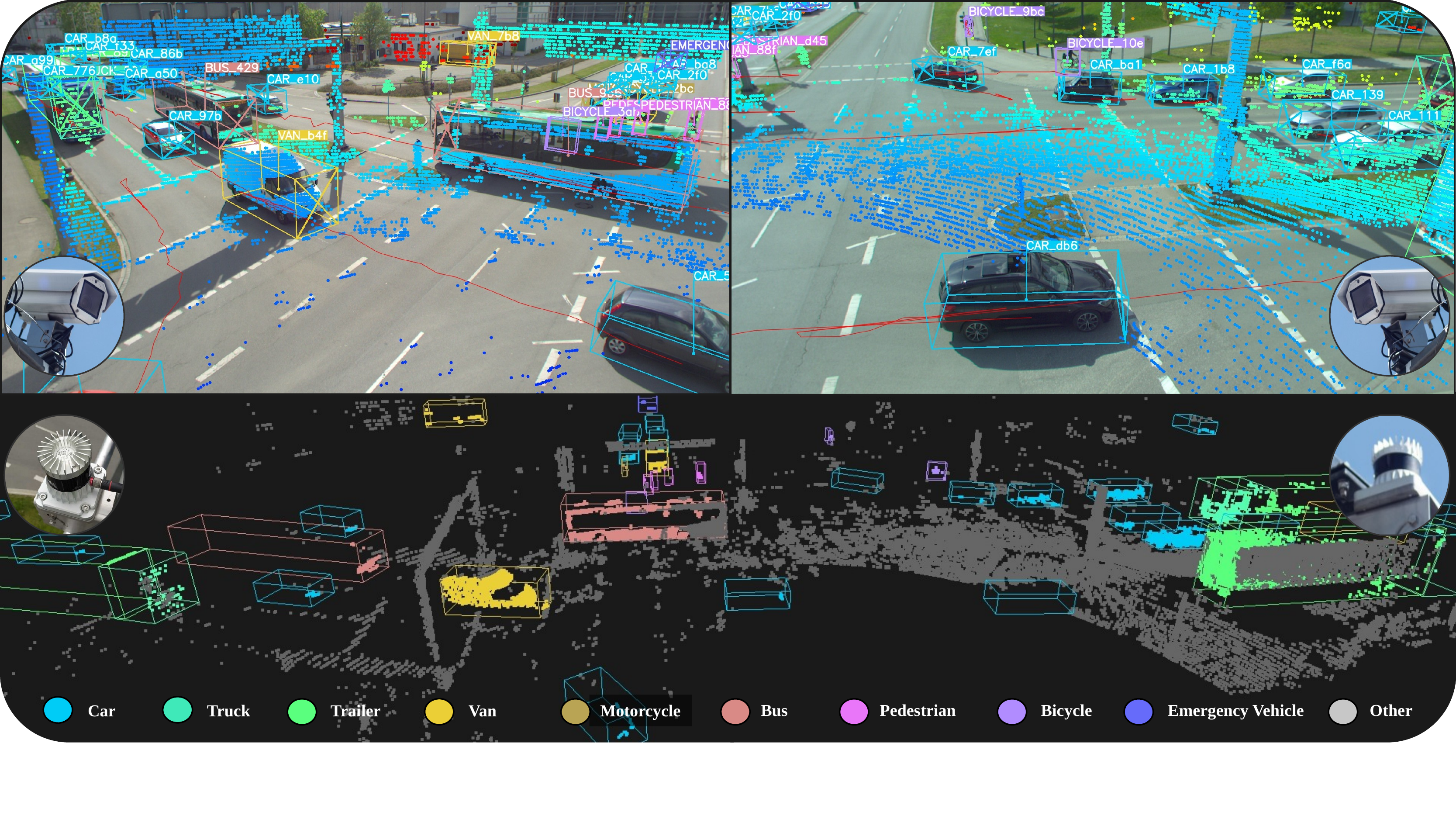}\bigskip
		\label{fig:overview_figure}
		\captionof{figure}{Visualization of 3D box labels and tracks in the A9 Intersection Dataset. The first row shows the labels projected into the two camera images. Below a registered point cloud from two LiDARs contains 3D box labels of the same scene.}
	}
	
	\makeatother

	\maketitle
	
	\thispagestyle{empty}
	\pagestyle{empty}
	
	\begin{abstract}
		Intelligent Transportation Systems (ITS) allow a drastic expansion of the visibility range and decrease occlusions for autonomous driving. To obtain accurate detections, detailed labeled sensor data for training is required. Unfortunately, high-quality 3D labels of LiDAR point clouds from the infrastructure perspective of an intersection are still rare. Therefore, we provide the A9 Intersection Dataset, which consists of labeled LiDAR point clouds and synchronized camera images. Here, we recorded the sensor output from two roadside cameras and LiDARs mounted on intersection gantry bridges. The point clouds were labeled in 3D by experienced annotators. Furthermore, we provide calibration data between all sensors, which allow the projection of the 3D labels into the camera images and an accurate data fusion. Our dataset consists of 4.8k images and point clouds with more than 57.4k manually labeled 3D boxes. With ten object classes, it has a high diversity of road users in complex driving maneuvers, such as left and right turns, overtaking, and U-turns. In experiments, we provided multiple baselines for the perception tasks. Overall, our dataset is a valuable contribution to the scientific community to perform complex 3D camera-LiDAR roadside perception tasks. Find data, code, and more information at \href{https://a9-dataset.com}{https://a9-dataset.com}.
	\end{abstract}
	
	
	
\newcolumntype{N}{@{}m{0pt}@{}}

\begin{table*}[t!]
  \caption{Comparison of popular autonomous driving datasets. Here we compare the perspective, the number of frames, the number of classes, the number of labeled objects, the number of tracks, and the license terms. The datasets cover different view perspectives: the vehicle view (V), from the steep elevated view (SE), and from the roadside view (R).\label{tbl:perception_datasets} }
  \centering
  \begin{tabular*}{\textwidth}{@{\extracolsep{\fill}}lccccccccN}
    \hline
    Name & Year & Perspective & \# Point Clouds &  \# Images & \# Classes & \# 3D Labels & \# Tracks & License \\
    \hline
    KITTI \cite{geiger2013vision}       & 2013 & V & 15.4k  & 15k   & 8     & 80k & - & CC BY-NC-SA 3.0\\
    Cityscapes \cite{MariusCordts.}     & 2016 & V & -      & 25k   & \textbf{30}    & - & - & Non-Commercial Use \\
    highD \cite{Krajewski.2018}         & 2018 & SE & -      & \underline{1.4M}  & 2     & - & 20k & Custom \\
    nuScenes\cite{caesar2020nuscenes}   & 2020 & V & \textbf{400k}   & \underline{1.4M}  & \underline{23}   & 1.4M  & - & CC BY-NC-SA 4.0\\
    Waymo \cite{sun2020scalability}     & 2020 & V & \underline{200}   & 1M    & 4     & \textbf{12.6M} & \textbf{7.6M} & Custom\\
    inD \cite{bock2020ind}              & 2020 & SE & -      & 0.9M  & 5     & - & 11.5k & Custom \\
    rounD \cite{Krajewski.2020}        & 2020 & SE & -      & 0.5M  & 8     & - & 13.7k & Custom\\
    exiD \cite{Moers.2022}              & 2022 & SE & -      & 1.4M  & 3     & - & 69k & Custom \\
    MONA \cite{gressenbuch2022mona}     & 2022 & SE & -      & \textbf{11.7M} & 2     & -  & \underline{702k} & Custom \\
    DAIR-V2X$^{\mathrm{*}}$ \cite{yu2022dair}          & 2022 & V,R& 71k    & 71k   & 10    & 1.2M & - & Non-Commercial Use\\
    IPS300+$^{\mathrm{**}}$ \cite{wang2022ips300}       & 2022 & R & 14.1k  & 14.1k & 7     & \underline{4.5M}  & - & CC BY-NC-SA 4.0\\
    Rope3D \cite{ye2022rope3d}          & 2022 & R & -      & 50k   & 13    & 1.5M  & - & Non-Commercial Use\\
    LUMPI \cite{busch2022lumpi}         & 2022 & R & 90k    & 200k  & 6     & - & - & CC BY-NC 3.0\\
    A9                                  & 2022 & R & 5.3k   & 5.4k  & 10    & 71.9k & 506 & CC BY-NC-ND 4.0\\
    - A9 \cite{cress2022a9}     & 2022 & R & 0.5k   & 0.6k  & 9     & 14.5k   & - & CC BY-NC-ND 4.0\\
    - \textbf{A9-I (Ours)}   & 2023 & R & 4.8k   & 4.8k  & 10    & 57.4k  & 506  & CC BY-NC-ND 4.0\\
    \hline\\[-8pt]
    \multicolumn{9}{l}{$^{\mathrm{*}}$~40\% of data from roadside perspective, 60\% from vehicle perspective.} \\
    \multicolumn{9}{l}{$^{\mathrm{**}}$Trucks and buses are sparsely represented which can lead to a limited perception performance.} \\
  \end{tabular*}
\end{table*}

\section{INTRODUCTION}
The roadside deployment of high-tech sensors to detect road traffic participants offers significant added value for intelligent and autonomous driving. This technology allows the vehicle to react to events and situations that are not covered by the vehicle's internal sensor range. Thus, the advantage is the drastic expansion of the field of view and the reduction of occlusions. For this reason, we can observe a continuous increase in Intelligent Transportation Systems (ITS) world-wide. It is noticeable that cameras and increasingly LiDARS are used to create a live digital twin of road traffic \cite{Cre.10122021}. To obtain accurate detections with such sensor systems, labeled sensor data is required for training.

Numerous datasets in the field of intelligent and autonomous driving have already been created. Datasets like \cite{geiger2013vision,caesar2020nuscenes,MariusCordts.,sun2020scalability} are taken from the vehicle perspective. In contrast, \cite{Krajewski.2018,Bock.2020,Krajewski.2020,Moers.2022,gressenbuch2022mona} are recorded from a very steep elevated view from a drone or a high building, so they are more suitable for trajectory prediction and tracking tasks. They are less suitable for 3D object detection because vehicles are far away and are only observed from above. Recently, a few datasets \cite{yu2022dair,wang2022ips300,ye2022rope3d,busch2022lumpi, cress2022a9} have been acquired from an roadside perspective and are thus suitable for improving perception algorithms for ITS. However, some datasets have deficiencies in their labeling quality, which harm the training of the algorithms (e.g., censored image areas with filled rectangles), or they lack certain vehicle classes (e.g., missing trucks and buses), or the datasets are too small in terms of 3D box labels and attributes.

According to the work mentioned, it can be recognized that high-quality 3D box labels of LiDAR point clouds from the roadside perspective with a wide diversity of traffic participants and scenarios are still rare. Therefore, our A9 Intersection (A9-I) Dataset provides LiDAR point clouds and camera images from a road intersection. The $4.8\text{k}$ labeled point cloud frames, which were labeled by experts, contain complex driving maneuvers such as left and right turns, overtaking maneuvers, and U-turns. With its ten object classes, our dataset has a high variety of road users, including vulnerable road users. Furthermore, we provide synchronized camera images and the extrinsic calibration data between LiDARs and the cameras. These matrices allow the projection of the 3D box labels to the camera images. All in all, our A9-I offers synchronized $4.8\text{k}$ images and $4.8\text{k}$ point clouds with $57.4\text{k}$ 3D box labels with track IDs that were manually labeled. In this work, we show additional comprehensive statistics and the effectiveness of our dataset. Over and beyond, we would like to emphasize that A9-I is an extension of our previous debut the A9 Dataset \cite{cress2022a9}, which covers highway traffic scenarios. Thus, we extend the existing A9 Dataset with additional traffic scenarios on a crowdy intersection and scale it up from $15\text{k}$ labeled 3D box labels to $57.4\text{k}$ including vulnerable road users. In evaluation experiments, we provide multiple baselines for the 3D perception task of 3D object detection with a monocular camera, a LiDAR sensor, and a multi-modal camera-LiDAR setup. Last but not least, we offer our dataset in OpenLABEL format under the Creative Commons License CC BY-NC-ND 4.0 so that it can be widely used by the scientific research community.\\

In summary, our \textbf{contributions} are:
\begin{itemize}
  \item A detailed and diverse dataset of $4.8\text{k}$ camera images as well as $4.8\text{k}$ labeled LiDAR point cloud frames. Thereby, we used two synchronized cameras and LiDARS, which cover an intersection from an elevated view of an ITS.
  \item Extrinsic calibration data between cameras and LiDARs allow an early and late fusion of objects. \item We provide an extensive A9-Devkit to load, transform, split, evaluate and visualize the data.
  \item $57.4\text{k}$ high-quality manually labeled 3D boxes with $273\text{k}$ attributes for both LiDARS resulting in $38\text{k}$ 3D box labels after data fusion. 
  \item Comprehensive statistics and analysis of the labels, number of points, occlusions, and tracks on the dataset, and the distribution of ten different object classes of road traffic.
  \item Multiple baselines for the 3D perception task of 3D object detection with a monocular camera, a LiDAR sensor, and a multi-modal camera-LiDAR setup.
\end{itemize}

\section{RELATED WORK}

As part of the development in the field of autonomous driving and intelligent vehicles, the number of datasets is increasing rapidly. The most popular datasets in this field are KITTI \cite{geiger2013vision}, nuScenes \cite{caesar2020nuscenes}, Cityscapes \cite{MariusCordts.}, and Waymo Open dataset \cite{sun2020scalability}. Except for the Cityscapes, the datasets provide labeled camera images and LiDAR point clouds. These datasets are used to train perception algorithms. Unfortunately, these valuable datasets only contain data from a vehicle's perspective. Therefore, this ego perspective is suboptimal for transfer learning. Networks trained on a dataset from the vehicle's perspective do not perform well on data obtained, e.g. from a roadside perspective.

Another sensor perspective is, for example, the elevated view. With this, the scene can ideally be viewed without occlusions. To achieve a high level of perception for this elevated view, training with appropriate datasets is necessary. The focus of the drone dataset family highD \cite{Krajewski.2018}, inD \cite{Bock.2020}, rounD \cite{Krajewski.2020}, and exiD \cite{Moers.2022} is the trajectory of road users in the city as well as in the freeway area. The datasets were recorded by a drone and provide a vast top-down view of the scene. The main limitation is the limited recording time in challenging weather conditions. To overcome this drone-related issue, the MONA \cite{gressenbuch2022mona} dataset provides data that was created with a camera mounted on a building. On the one hand, these datasets are ideal for trajectory research, because they were recorded from a very steep angle to the road. On the other hand, they are less suitable for 3D object detection, because of the missing 3D dimensions.

A dataset, which contains data from an elevated view of an ITS with an angle that is not too steep, is the DAIR-V2X \cite{yu2022dair}. The main focus of DAIR-V2X is the support of of 3D object detection tasks. It consists of $71\text{k}$ labeled camera images and LiDAR point clouds, $40\%$ of which are from an roadside infrastructure. For this purpose, the dataset covers city roads, highways, and intersections in different weather and lighting conditions. Unfortunately, no exact statistics for this variation or exact sensor specifications are available. As a last point, the quality of the data is further compromised by filled rectangles over privacy-sensitive image areas (e.g., license plates), which can lead to problems during training for object detection. Another dataset from the roadside infrastructure perspective with a camera and LiDAR combination is the IPS300+ \cite{wang2022ips300}. The dataset includes $14\text{k}$ data frames, with an average of $319$ labels per frame. They used $1$ LiDAR and $2$ cameras as a stereo setup with a lens focal length of $4.57~\text{mm}$. The dataset was recorded several times a day at one intersection and provides seven different object categories: car, cyclist, pedestrian, tricycle, bus, truck, and engineer car. According to the statistics, unfortunately, there is less representation in the classes of trucks and buses, so that the recognition of these classes will probably be poor. The Roadside Perception 3D dataset (Rope3D) \cite{ye2022rope3d} provides $50\text{k}$ images including 3D box labels from an monocular infrastructure camera at an intersection. The missing 3D information of the detected objects in the 2D camera image was added with a LiDAR, which was mounted on a vehicle. In total, the images contain over $1.5\text{M}$ labeled 3D boxes, $670\text{k}$ 2D bounding boxes, in various scenes at different times (daytime, night, dawn/dusk), different weather conditions (sunny, cloudy, rainy), and different traffic densities. Furthermore, the objects are divided into $13$ classes with several attributes. Another roadside infrastructure dataset is LUMPI \cite{busch2022lumpi}, which was recorded at an intersection in Hanover, Germany. For this purpose, a total of $200\text{k}$ images as well as $90\text{k}$ point clouds were acquired. Three different cameras and five different LiDARs provide several field of views on the scene. Here, different sensor configurations were used for the recordings. The sensor perspective is from a vehicle as well as from the roadside infrastructure. Unfortunately, the number of labels and other detailed information about the labeled objects was not provided. A further contribution in the field of roadside infrastructure data for training perception algorithm is the A9-Dataset \cite{cress2022a9}. It is our preliminary work and includes $642$ camera images and $456$ LiDAR point clouds. In total, this dataset consists of 1k 3D box labels. The charm is that most camera images contain the same traffic scene from four different viewpoints. Here, we labeled 14k 3D boxes. Moreover, the frames contain $13.17$ 3D box labels in average. For this purpose, we supported the common classes of car, trailer, truck, van, pedestrian, bicycle, bus, and motorcycle in the domain of a highway. The main limitations in our previous work were firstly the small number of labeled LiDAR point clouds and secondly that we only had a simple highway scenario. For this reason, we present an extension to our dataset that addresses these weaknesses.

\begin{figure}[t!]
	\centering
	\includegraphics[width=1\linewidth]{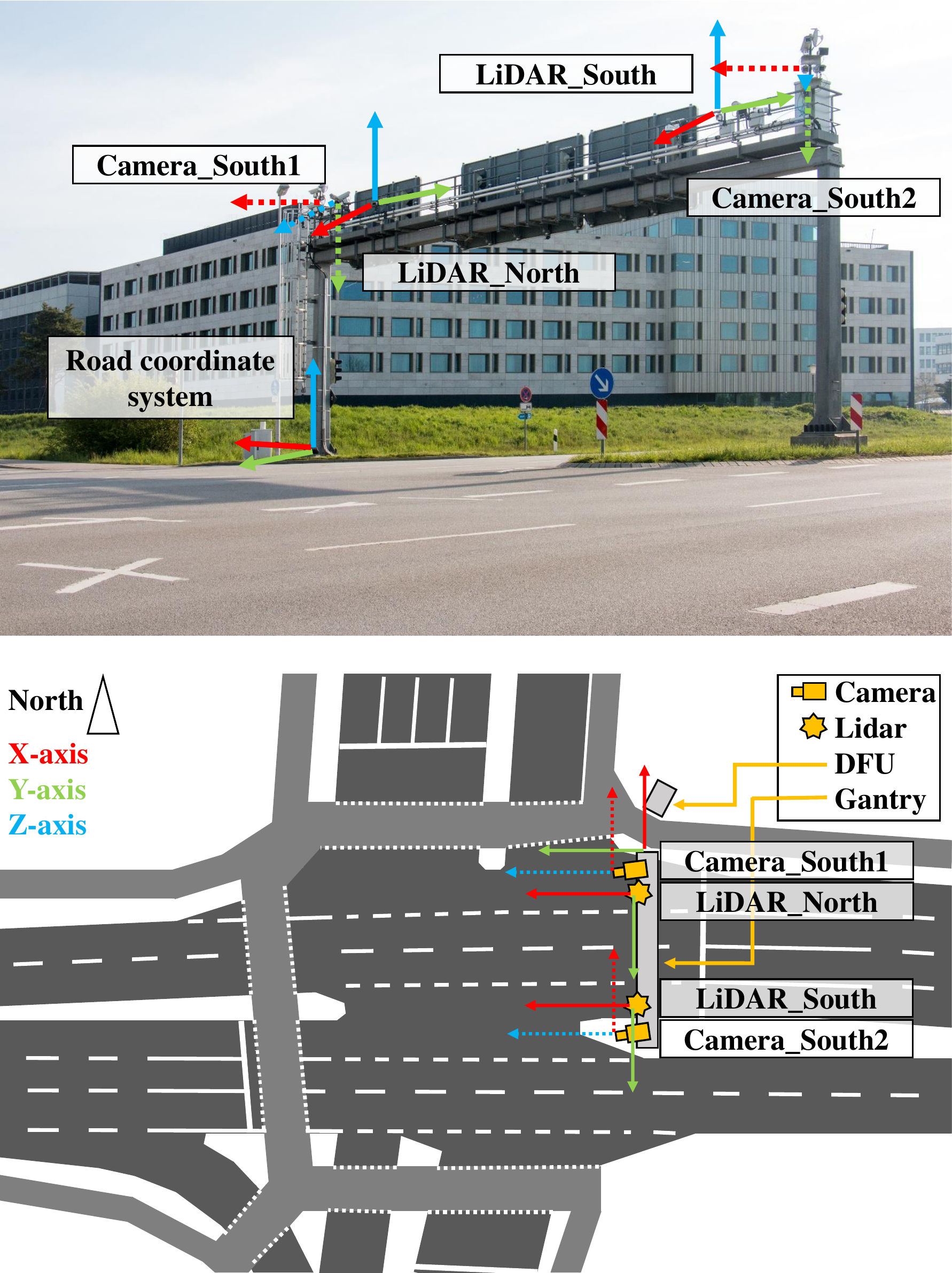}
	\caption{Two cameras and two LiDARs are used to create the A9 Intersection Dataset. The sensors are mounted on a sign gantry and thus record the central intersection area. Then, the Data Fusion Unit (DFU) process the data streams, which results in a fused digital twin of the road traffic. Furthermore, the coordinate systems of the individual sensors and the road coordinate system, which was defined at the northern stem of the bridge, can be taken from the figure.} 
	\label{fig:intersection_sketch}
\end{figure}

\section{A9 INTERSECTION DATASET}
In this section, we present the A9 Intersection Dataset. It is an extension of our previous work, the A9-Dataset \cite{cress2022a9}, which covers the highway domain. We describe the sensor setup at our intersection, the data selection and annotation process, and the data structure used. Last, this section contains comprehensive statistics and an introduction to our A9-Devkit.
\subsection{Sensor Setup}
The A9-I Dataset is recorded on the ITS testbed, which was established as part of the Providentia++ project \cite{krammer2022providentia, providentia-webpage}. Here, roadside sensors are set up on a gantry located at the intersection of Schleißheimer Straße (B471) and Zeppelinstraße in Garching near Munich, Germany. For this dataset, we use two cameras and two LiDARs with the following specifications:
\begin{itemize}
  \item  \textbf{Camera:} Basler ace acA1920-50gc, $1920\times1200$, Sony IMX174, glo. shutter, color, GigE with 8 mm lenses.
  \item  \textbf{LiDAR:} Ouster OS1-64 (gen. 2), 64 vert. layers, $360 \text{° FOV}$, below horizon configuration, $120~\text{m}$ range, $1.5-10~\text{cm}$ accuracy.
\end{itemize}
The sensors are mounted side by side on the gantry, as shown in Figure \ref{fig:intersection_sketch}. Here, the sensors detect the traffic in the center of the intersection from a height of $7~\text{m}$. It is worth mentioning that the cameras and LiDARs are spatiotemporally calibrated. For the temporal calibration, we synchronized the sensors with a Network Time Protocol (NTP) time server, for the extrinsic calibration between the cameras and the LiDARs, we used a targetless extrinsic calibration method, which was inspired by \cite{chongjian2021pixel}.

\subsection{Data Selection and Annotation}
We select the data based on interesting and challenging traffic scenarios like left, right, and U-turns, overtaking maneuvers, tail-gate events, and lane merge scenarios. Furthermore, we take highly diverse and dense traffic situations into account, so that we get an average of over $15$ road users per frame. To cover diverse weather and light conditions in our A9-I Dataset, it consists of $25\%$ nighttime data including heavy rain, and $75\%$ daytime data with sunny and cloudy weather conditions. This enables a good performance of the detector even in challenging weather conditions.

We record camera data at $25 \text{ Hz}$ and LiDAR data at $10 \text{ Hz}$ into \textit{rosbag} files. Then we extract the raw data and synchronize the camera and LiDAR frames at $10 \text{Hz}$ based on timestamps. Based on the raw data of the LiDAR point clouds, 3D box labels were created by experts. As all four sensors are cross-calibrated, we can also use these 3D box labels from the point cloud to evaluate monocular 3D object detection algorithms. Since the labeling quality of the test sequence is very important, it was reviewed multiple times by us. Here, we improve the labeling quality by using our preliminary \textit{proAnno} labeling toolbox \cite{zimmer20193d}.

\subsection{Data Structure}
Our dataset is divided into subsets \textit{S1} through \textit{S4}, which contain continuous camera and labeled LiDAR recordings. Set \textit{S1} and \textit{S2} are each 30 seconds long and demonstrate a daytime scenario at dusk. A 120-second long sequence during daytime and sunshine can be found in sequence \textit{S3}. Sequence \textit{S4} contains 30-second data recording at night and in heavy rain. The file structure is given below:
{\scriptsize{

\dirtree{%
.1 a9-intersection-dataset. 
.2 a9\_dataset\_r02\_s01.
.3 {point\textunderscore clouds}.
.4 {s110\textunderscore lidar\textunderscore ouster\textunderscore north}.
.5 {timestamp\textunderscore sensor\textunderscore id.pcd}.
.4 {s110\textunderscore lidar\textunderscore ouster\textunderscore south}.
.5 {timestamp\textunderscore sensor\textunderscore id.pcd}.
.3 images.
.4 {s110\textunderscore camera\textunderscore basler\textunderscore south1\textunderscore8mm}.
.5 {timestamp\textunderscore sensor\textunderscore id.jpg}.
.4 {s110\textunderscore camera\textunderscore basler\textunderscore south2\textunderscore8mm}.
.5 {timestamp\textunderscore sensor\textunderscore id.jpg}.
.3 {labels}.
.4 {s110\textunderscore lidar\textunderscore ouster\textunderscore north}.
.5 {timestamp\textunderscore sensor\textunderscore id.json}.
.4 {s110\textunderscore lidar\textunderscore ouster\textunderscore south}.
.5 {timestamp\textunderscore sensor\textunderscore id.json}.
.2 a9\_dataset\_r02\_s02.
.2 a9\_dataset\_r02\_s03.
.2 a9\_dataset\_r02\_s04.
}
}}

All labeled data is in \textit{OpenLABEL} format \cite{hagedorn2021open}. \textit{OpenLABEL} files are stored in \textit{.json} format. One file contains all labeled objects of a single frame with 32-bit long unique identifiers (UUIDs), the position, dimensions, rotation, and the attributes like the occlusion level, the body color, the number of trailers, the specific object type, and the number of 3D points. Furthermore, a frame contains properties like the exact epoch timestamp, the weather type, the time of day, and the corresponding image and point cloud file names. In \textit{OpenLABEL} the label files also contain the calibration data -- intrinsic and extrinsic information. 

\begin{figure*}[h!] 
	\centering
	\subfloat[Object classes including occlusion level. \label{fig:bar_chart_occlusion_level}]{%
		\includegraphics[width=0.33\linewidth]{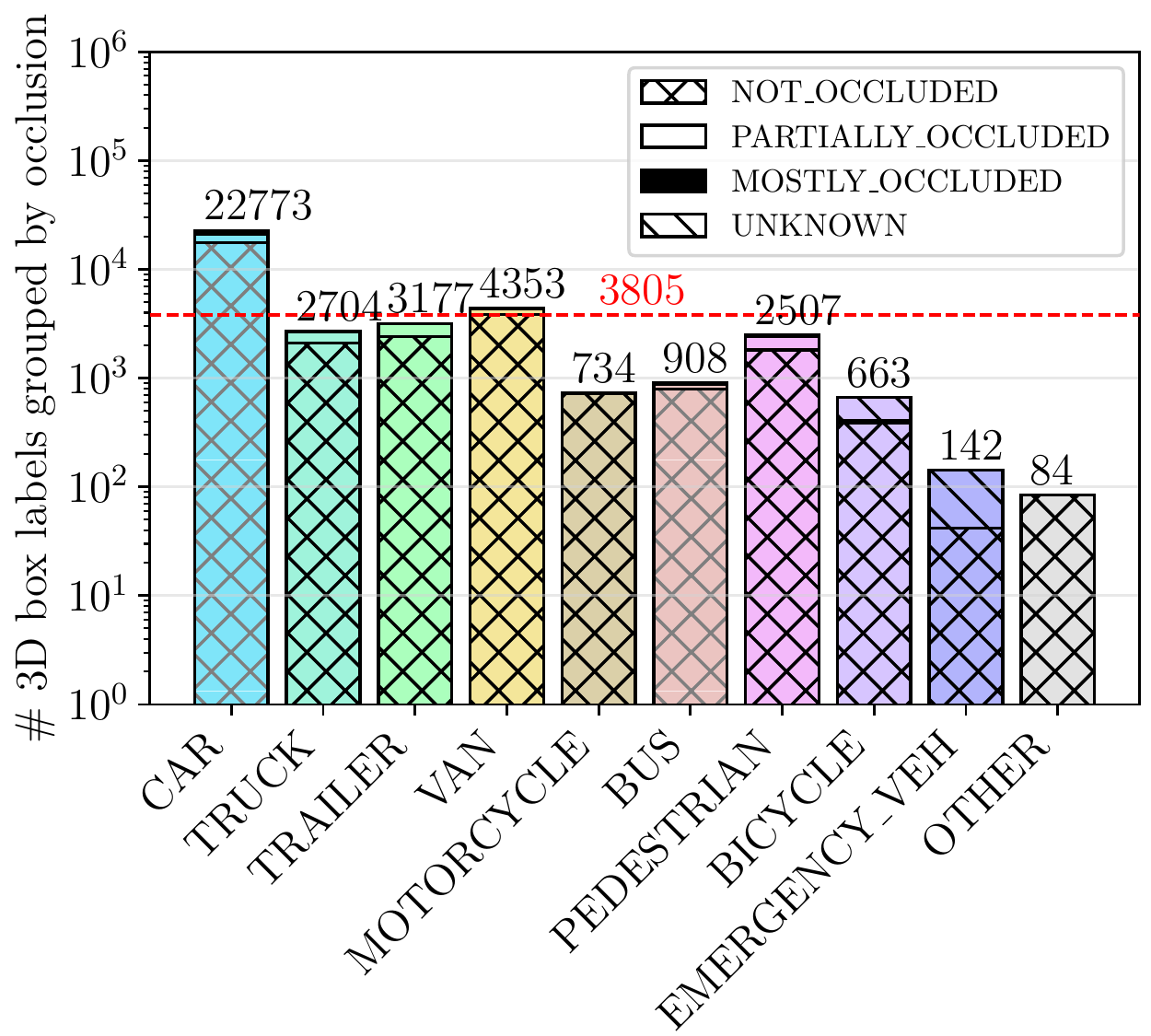}}
	\hfill
	\subfloat[Labeled 3D box labels per frame. \label{fig:histogram_objects_in_frame}]{%
		\includegraphics[width=0.33\linewidth]{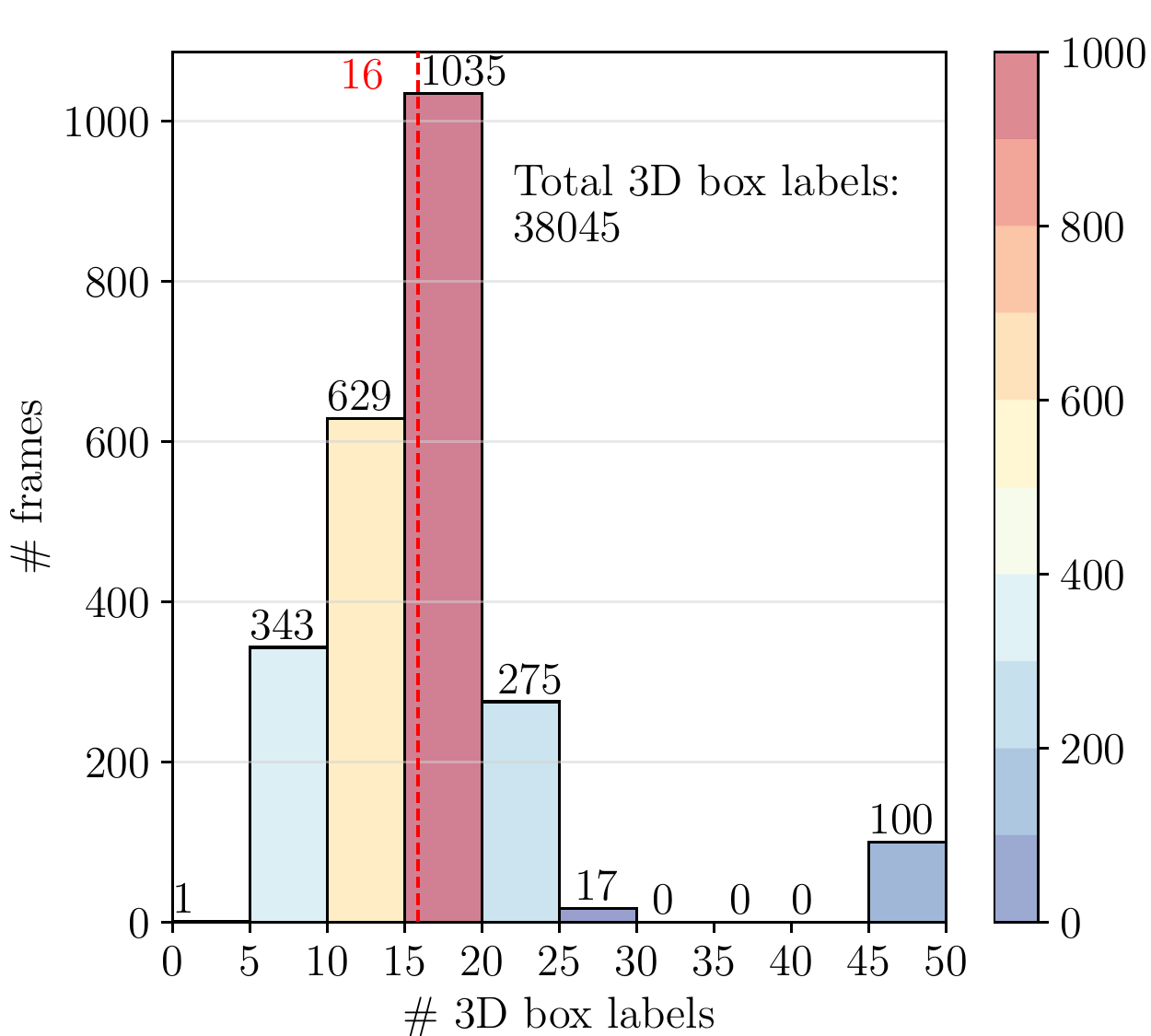}}
	\hfill
	\subfloat[Rotations of 3D box labels. \label{fig:histogram_avg_rotation}]{%
		\includegraphics[width=0.33\linewidth]{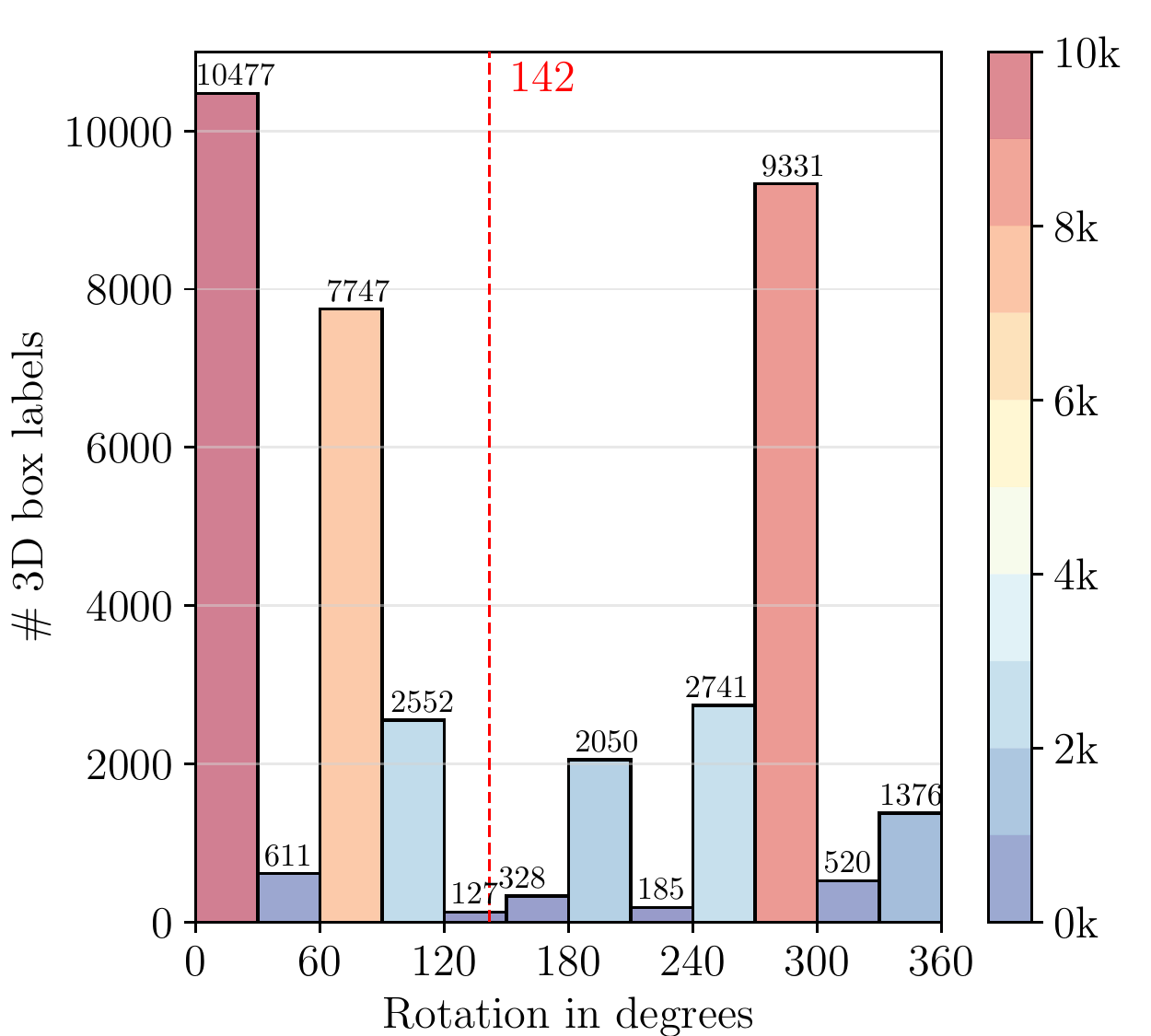}}
	\caption{The A9-I Dataset consists of $38,045$ 3D box labels after data fusion, where the \textit{CAR} class is dominant and $78.2\%$ of the objects are occlusion-free. The 3D box labels show different rotations, which is due to the road traffic in an intersection.}
	\label{fig:statistic_labels}
\end{figure*}
\begin{figure*}[h!]
	\centering
	\subfloat[Average number of points per class. \label{fig:bar_chart_avg_num_points_per_class}]{%
		\includegraphics[width=0.33\linewidth]{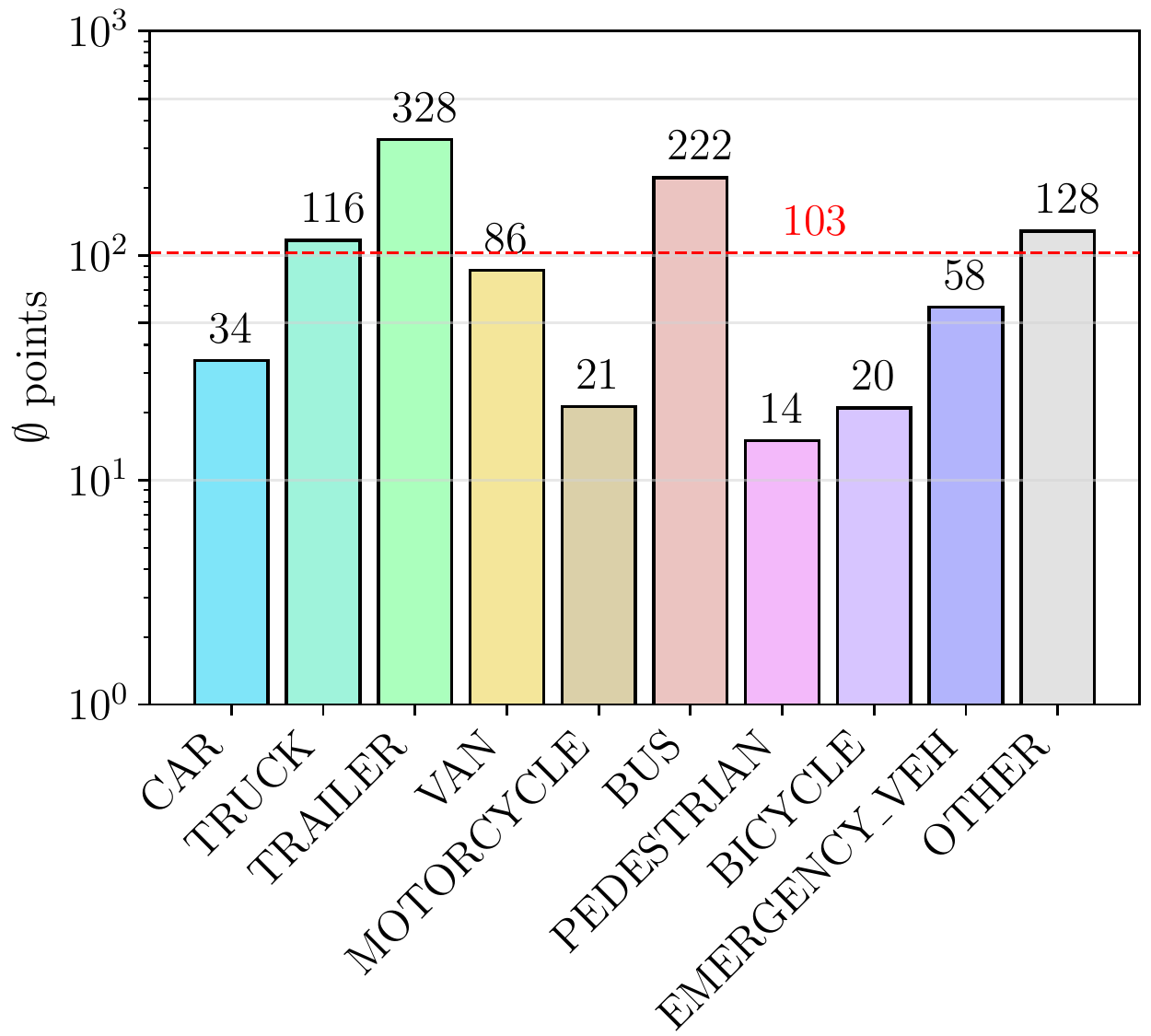}}
	\hfill
	\subfloat[Number of points grouped by distance. \label{fig:line_chart_avg_num_points_with_distance}]{%
		\includegraphics[width=0.33\linewidth]{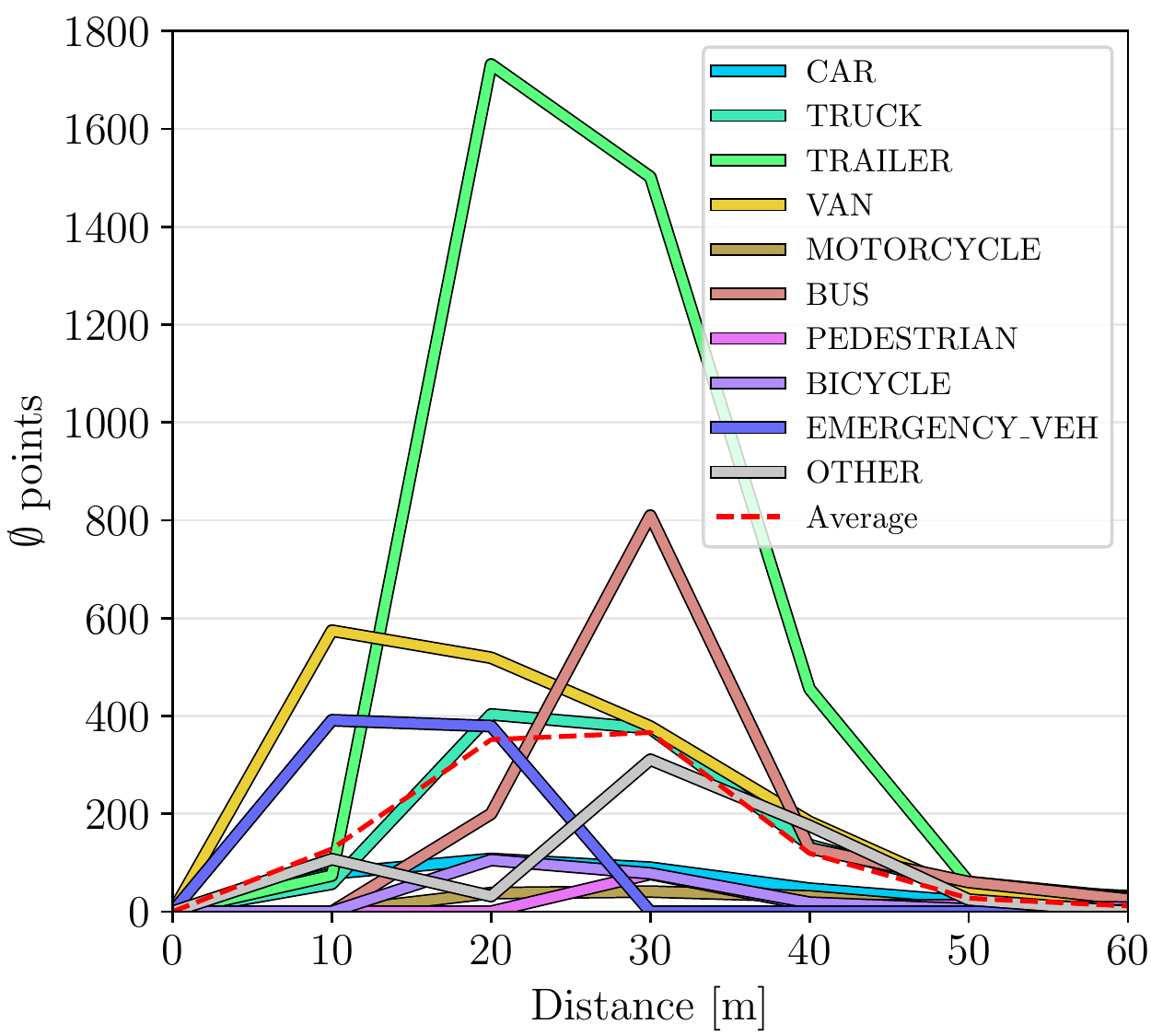}}
	\hfill
	\subfloat[Number of points in a 3D box label. \label{fig:histogram_num_objects_per_num_points}]{%
		\includegraphics[width=0.33\linewidth]{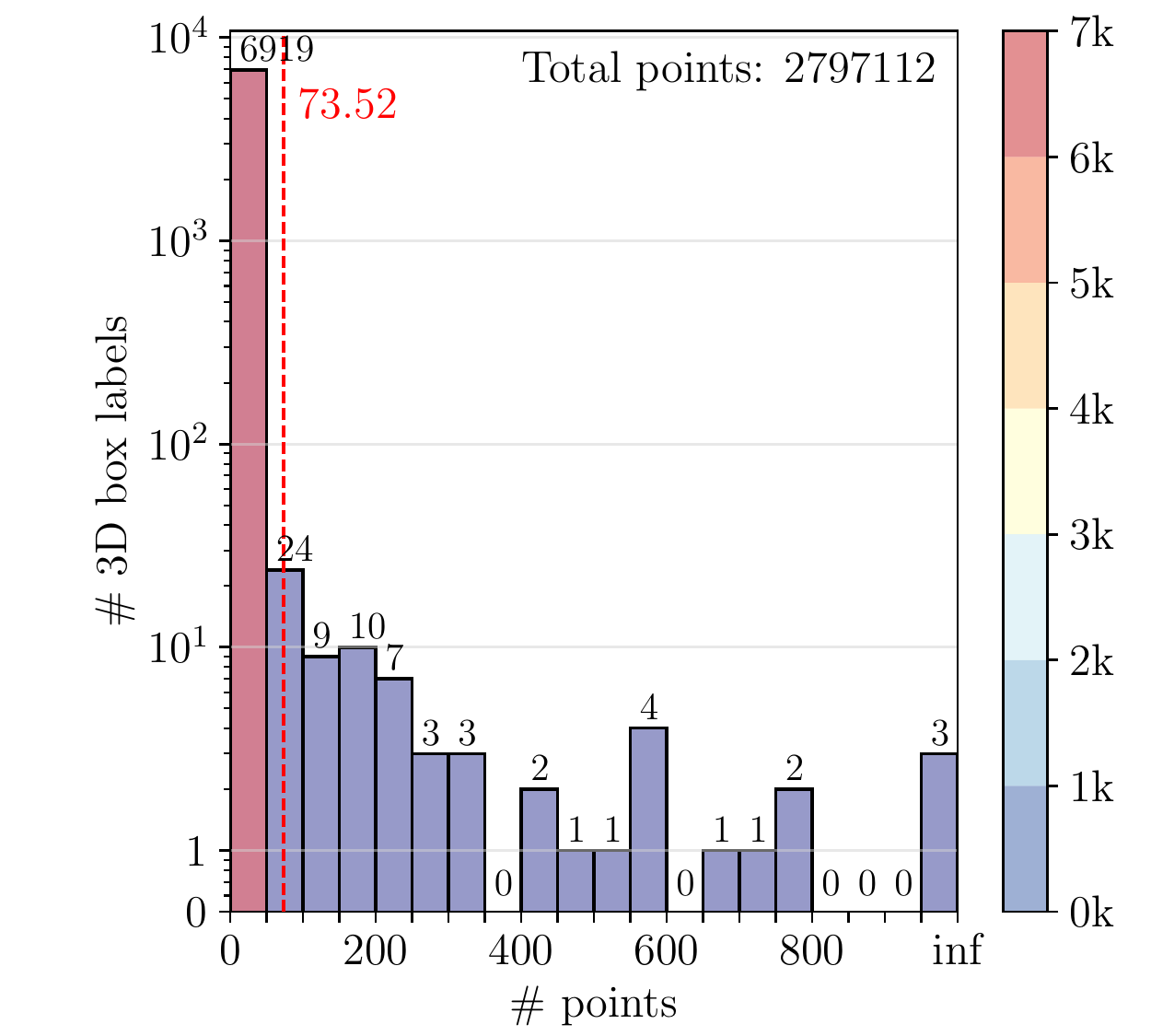}}
	\caption{As expected, there is a causality between the dimensions of road users and the number of points. Most classes have the highest number of points at a distance between $10$ to $30$ meters. On average, each 3D box label contains $73.52$ points.}
	\label{fig:statistic_points} 
\end{figure*}

\begin{figure*}[h!]
    \centering
	\subfloat[Average track length per class. \label{fig:bar_chart_avg_track_length}]{%
		\includegraphics[width=0.33\linewidth]{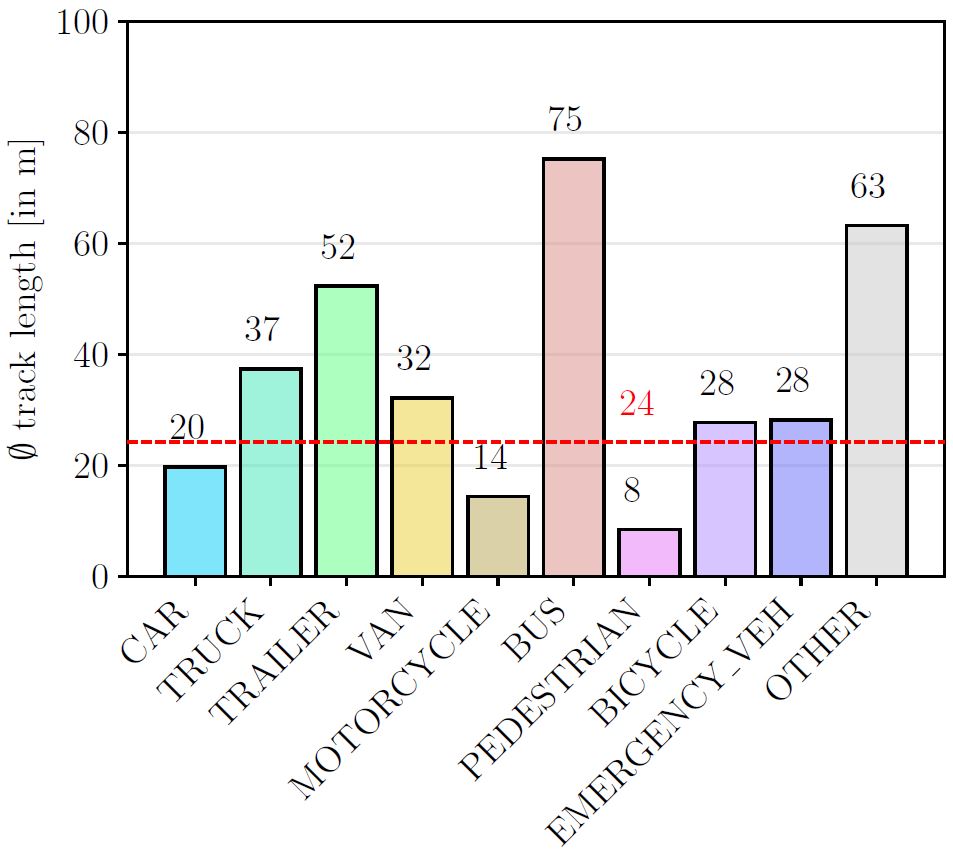}}
	\hfill
	\subfloat[Distribution of the several track lengths. \label{fig:histogram_track_lengths}]{%
		\includegraphics[width=0.33\linewidth]{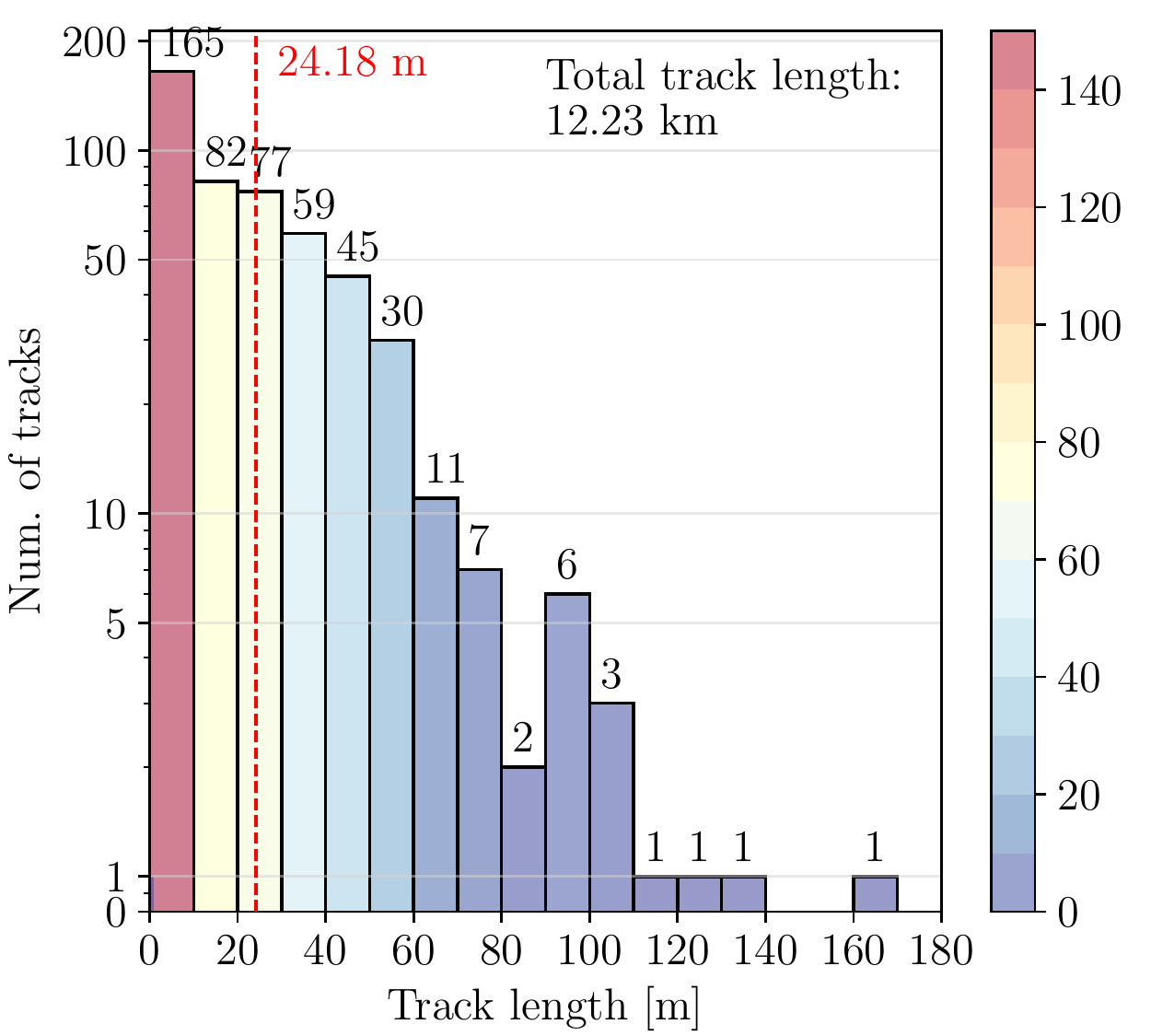}}
	\hfill
	\subfloat[Visualization of tracks in BEV. \label{fig:visualization_tracks}]{%
		\includegraphics[height=5.1cm,width=0.33\linewidth]{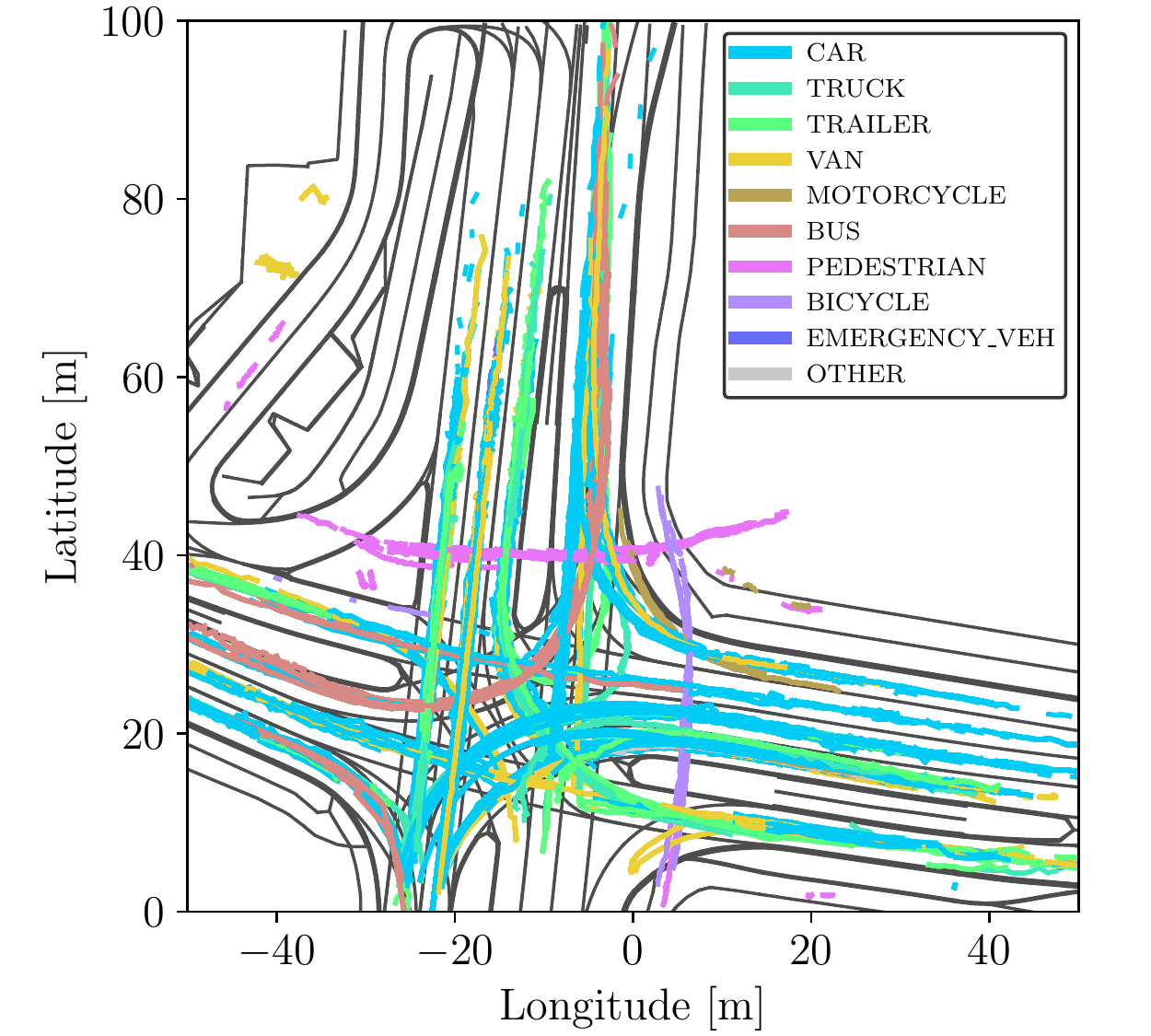}}
	\caption{The 3D box label of each traffic participant has always received the same tracking ID in successive frames in a sequence. The average track length is $24.18~\text{m}$. In total, our 3D box labels have a track length of $12.23~\text{km}$.}	
	\label{fig:statistic_tracks} 
\end{figure*}

We suggest a split into training ($80\%$), validation ($10\%$), and test set ($10\%$). The test set is made up of a continuous sequence with track IDs, as well as randomly sampled frames from four different scenarios and daytimes. We sample frames using stratified sampling to create a balanced dataset among sensor types, weather scenarios, and day times. To prevent overfitting, we do not publish our test set labels.

\subsection{Data Statistics}

In total, we provide $4,800$ labeled LiDAR point cloud frames sampled from four different sequences. Here, $57,406$ 3D objects ($506$ unique objects) were annotated with $273,861$ object attributes. After fusing the labels from both LiDARs we get $38,045$ registered 3D objects ($482$ unique objects) with $171,045$ attributes. The following statistics refer to the fusion result with the complete dataset inclusive of training, validation, and test set. In \Cref{tbl:object_classes}, we can see an overview of the registered 3D box labels.
\begin{table}[h!]
	\caption{The total number of labels, average dimensions in meters, and the average number of 3D LiDAR points among all classes.}
	\label{tbl:object_classes}
	\centering
	\begin{tabular}{p{1.3cm}|rrrrr}
	    \hline
		Class & \#Labels & $\diameter$Length & $\diameter$Width & $\diameter$Height & $\diameter$Points\\
		\hline
		Car           & \textbf{22,773} & 4.27  & 1.91 & 1.59 & 34.03\\
		Truck         & 2,704  & 3.11  & 2.90 & \underline{3.43} & 116.87\\
		Trailer       & 3,177  & \underline{10.19} & \textbf{3.12} & \textbf{3.65} & \textbf{328.36}\\
		Van           & \underline{4,353}  & 6.35  & 2.52 & 2.47 & 86.11\\
		Motorcycle    & 734    & 1.90  & 0.83 & 1.60 & 21.23\\
		Bus           & 908    & \textbf{12.65} & \underline{2.95} & 3.27 & \underline{222.36}\\
		Pedestrian    & 2,507  & 0.80  & 0.73 & 1.72 & 14.98\\
		Bicycle       & 663    & 1.57  & 0.74 & 1.72 & 20.95\\
		Emergency Vehicle & 142& 6.72  & 2.35 & 2.35 & 58.95\\
		Other         & 84     & 5.28  & 1.92 & 1.90 & 128.17\\
		\hline
		Total         & 38,045 & -  & - & - & 103.20\\
		\hline
	\end{tabular}
\end{table}

A deep dive into distribution of the labels of our A9-I Dataset is provided in Figure \ref{fig:statistic_labels}. Here, the distribution of the ten object classes is shown. The vehicle class \textit{CAR} is dominant, followed by the classes \textit{TRUCK}, \textit{TRAILER}, \textit{VAN}, and \textit{PEDESTRIAN}, which occur in roughly the same order of magnitude. The classes \textit{MOTORCYCLE}, \textit{BUS}, \textit{BICYCLE}, \textit{EMERGENCY VEHICLES}, and \textit{OTHER} are present in a slightly smaller number. Since we have annotated the occlusion level for each 3D box label, we come to the result that $78.2\%$ were classified as \textit{NOT\_OCCLUDED}, $16.1\%$ as \textit{PARTIALLY\_OCCLUDED}, $0.8\%$ as \textit{MOSTLY\_OCCLUDED}, and $4.9\%$ were classified as \textit{UNKNOWN} (not labeled). It can also be seen that most of the labeled frames contain between $15$ and $20$ labeled 3D boxes. In 100 frames, there are even between $45$ and $50$ labeled 3D objects. Furthermore, the A9-I includes significantly more variations in the maneuvers of road users at the intersection, as compared to our previous work \cite{cress2022a9}. We can see three peaks where vehicles are moving to the south, north and east direction of the intersection. Vehicles moving between south and north are indicated by the peaks around $90$ and $270$ degrees. The smaller peaks adjacent to the main peaks correspond to turning maneuvers, such as right or left turns.

The labels are based on the LiDAR point clouds. In Figure \ref{fig:statistic_points}, we performed a detailed analysis of the points concerning the labeled classes, of the individual distances of the points concerning the labeled classes, and of the distribution of the points. Firstly, as expected, the correlation between the average number of points and the average size of the class can be observed. Here, the \textit{TRAILER} class, which has the highest height, also has the highest average number of points, followed by the \textit{BUS} class, which is the longest. Conversely, the \textit{PEDESTRIAN} class, which has the smallest size, has the lowest average number of points. Second, in general, due to the elevated position of the LiDARs, the field of view only starts to have an effect from about $10~\text{m}$ onwards. Most classes have the highest number of points at a distance between $10~\text{m}$ to $30~\text{m}$. Interestingly, the class \textit{TRAILER} has the highest average number of points at a distance between $15~\text{m}$ and $20~\text{m}$. With increasing distance, the average number of points is naturally declining. Lastly, all 3D box labels have a total of $2,797,112$ points. According to the distribution of the number of points per 3D box label, most of the boxes have a maximum of about $50$ points. However, the 3D box labels have on average $73.52$ points per object. 

In addition to the statistics about the labels and the underlying point clouds, we also analysed the calculated tracks, see Figure \ref{fig:statistic_tracks}. We were able to determine these trivially since the same tracking ID was selected for each consecutive frame when marking the 3D box labels. The average track length in our A9-I Dataset is $24.18~\text{m}$. Here, the class \textit{BUS} is very dominant with an average track length of $75~\text{m}$. The reason for this is because, firstly, the buses are very visible, and secondly, completely cross the intersection. All in all, the full dataset contains $506$ unique objects (3D box labels) with a total track length of $12.23~\text{km}$ with a maximum track length of $162.87~\text{m}$. Thus, our A9 Intersection Dataset can also be used to handle issues regarding tracking that are addressed by \cite{Strand.742022772022}.

\subsection{A9-Devkit}
To work with our A9-I Dataset, we provide the A9 Development Kit: \href{https://github.com/providentia-project/a9-dev-kit}{https://github.com/providentia-project/a9-dev-kit}. It contains a dataset loader to load images, point clouds, labels, and calibration data. Furthermore, we provide a converter from \textit{OpenLABEL} to multiple different dataset formats like \textit{KITTI}, \textit{COCO}, \textit{YOLO}, and the other way round. We follow the \textit{.json}-based \textit{OpenLABEL} standard \cite{hagedorn2021open} from the ASAM organization for the label structure. Some pre-processing scripts transform and filter the raw point cloud \textit{.pcd} \textit{ASCII} data into binary data to reduce the file size and make it compatible with point cloud loaders. In addition, a point cloud registration module can merge multiple point clouds to increase the point density. Finally, we provide a data visualization module to project the point clouds and labels into the camera images.

	\section{EVALUATION}

\begin{table*}[t]
    \centering
    \caption{Evaluation results on the A9-I Dataset test set (N=North, S=South, EF=Early Fusion, LF=Late Fusion). \\\hspace{\textwidth}We report the $mAP_{3D@0.1}$ results for the following six classes: \textit{Car}, \textit{Truck}, \textit{Bus}, \textit{Motorcycle}, \textit{Pedestrian}, \textit{Bicycle}. According to \cite{zimmer2023infra}, we crop the dataset into three subsets: A9-I-south1 (Camera\_South1), A9-I-south2 (Camera\_South2), and A9-I-full (Camera\_full). }
    \label{tbl:evaluation}
    \begin{tabular}{lllrrrr}
        \hline
        \textbf{FOV} & \textbf{Model}  & \textbf{Sensor Modality} & \multicolumn{4}{c}{$\boldsymbol{mAP_{3D}}$} \\
        &&&Easy& Mod. & Hard & Overall\\
         \hline
        Camera\_S1 & MonoDet3D\cite{zimmer2023infra}  & Camera\_S1         & 43.27 & 13.28 & 2.16 & 19.57\\
        Camera\_S1 & PointPillars$^{\mathrm{\star}}$\cite{lang2019pointpillars} & LiDAR\_N         & \textbf{76.19} & 34.58 & 30.00 & \underline{46.93}\\

        Camera\_S1 & PointPillars$^{\mathrm{\star}}$\cite{lang2019pointpillars} & LiDAR\_S         & 46.35 & \underline{41.05} & 24.16 & 37.18 \\
  
        Camera\_S1 & PointPillars$^{\mathrm{\star}}$\cite{lang2019pointpillars} & EF(LiDAR\_N + LiDAR\_S)   & \underline{75.81} & \textbf{47.66} & \textbf{42.16} & \textbf{55.21} \\
        Camera\_S1 & InfraDet3D\cite{zimmer2023infra} & LF(Camera\_S1 + EF(LiDAR\_N + LiDAR\_S))         & 67.08 & 31.38 & \underline{35.17} & 44.55 \\
                                                  \hline   
        Camera\_S2 & MonoDet3D\cite{zimmer2023infra} & Camera\_S2         & 16.82 & 27.87 & 26.67 & 23.78 \\

        Camera\_S2 & PointPillars$^{\mathrm{\star}}$\cite{lang2019pointpillars} & LiDAR\_N         & \underline{45.26} & 27.26 & 26.24 & 32.92 \\

        Camera\_S2 & PointPillars$^{\mathrm{\star}}$\cite{lang2019pointpillars} & LiDAR\_S        & 26.27 & \underline{38.24} & 13.16 & 25.89 \\

        Camera\_S2 & PointPillars$^{\mathrm{\star}}$\cite{lang2019pointpillars} & EF(LiDAR\_N + LiDAR\_S)   & 38.92 & \textbf{46.60} & \textbf{43.86} & \textbf{43.13}\\

        Camera\_S2 & InfraDet3D\cite{zimmer2023infra} & LF(Camera\_S2 + EF(LiDAR\_N + LiDAR\_S))         & \textbf{58.38} & 19.73 & \underline{33.08} & \underline{37.06}\\
                                                  \hline
        Camera\_full & MonoDet3D\cite{zimmer2023infra} & LF(Camera\_S1 + Camera\_S2)~~~~~~~~~~~~~~~~~~~~~~~~    &  19.05 & 24.12 & 25.55 & 22.91\\
        Camera\_full & PointPillars$^{\mathrm{\star}}$\cite{lang2019pointpillars} & LiDAR\_N        & \textbf{76.04} & 26.03 & 20.60  & \underline{40.89}\\

        Camera\_full & PointPillars$^{\mathrm{\star}}$\cite{lang2019pointpillars} & LiDAR\_S         &  38.82 & \underline{32.83} & 10.93 & 27.53\\
        Camera\_full & PointPillars$^{\mathrm{\star}}$\cite{lang2019pointpillars} & EF(LiDAR\_N + LiDAR\_S)   &  \underline{70.53} & \textbf{44.20} & \textbf{39.04} & \textbf{51.25}\\

        Camera\_full & InfraDet3D\cite{zimmer2023infra} & LF(LF(Camera\_S1+Camera\_S2) + LF(LiDAR\_N+LiDAR\_S))    & 47.27 & 26.15 & 19.71 & 31.04\\

        Camera\_full & InfraDet3D\cite{zimmer2023infra} & LF(LF(Camera\_S1+Camera\_S2) + EF(LiDAR\_N+LiDAR\_S))     & 64.30 & 23.83 & \underline{26.05} & 38.06\\
                                                  \hline\\[-8pt]
         \multicolumn{6}{l}{$^{\mathrm{\star}}$PointPillars inference score threshold is set to 0.3.} \\
         
    \end{tabular}
\end{table*}

\begin{figure*}[h!]
\centering
\minipage{0.33\textwidth}
  \includegraphics[width=\linewidth]{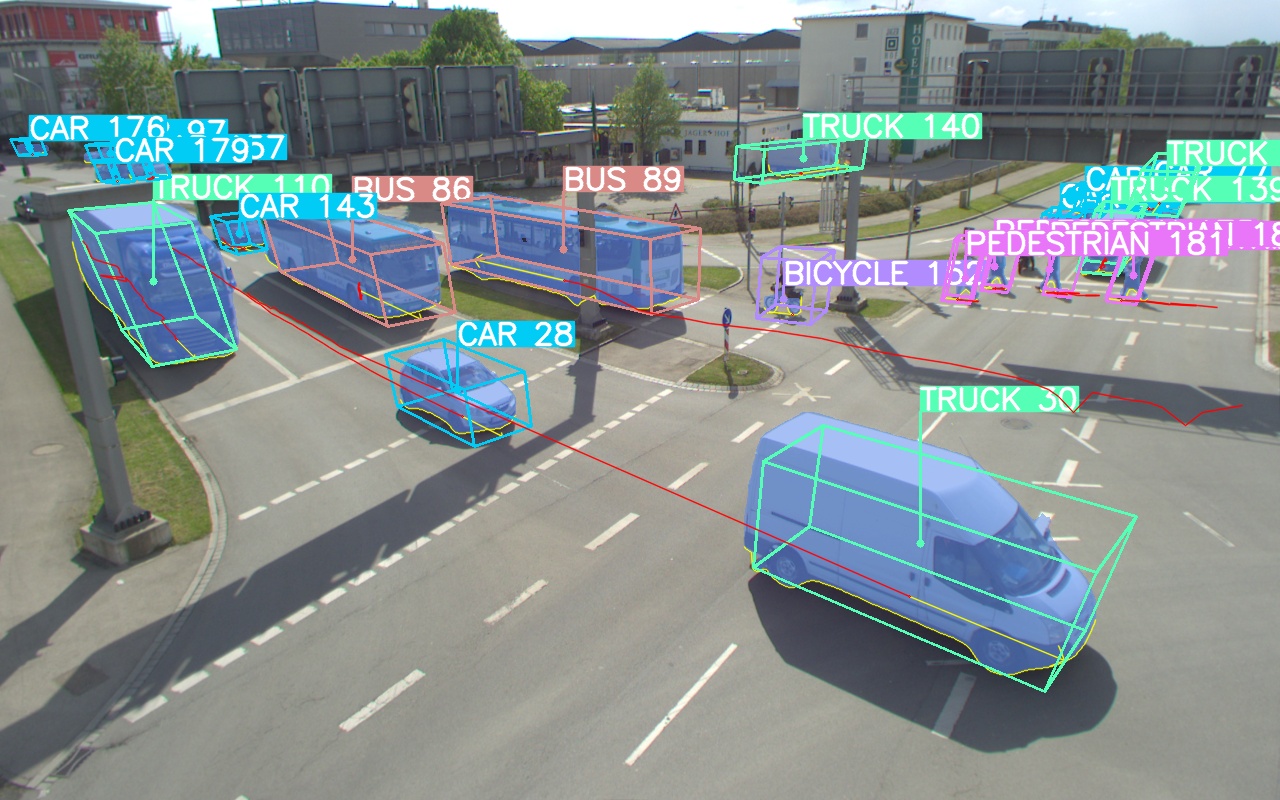}
\endminipage
\minipage{0.33\textwidth}
  \includegraphics[width=\linewidth]{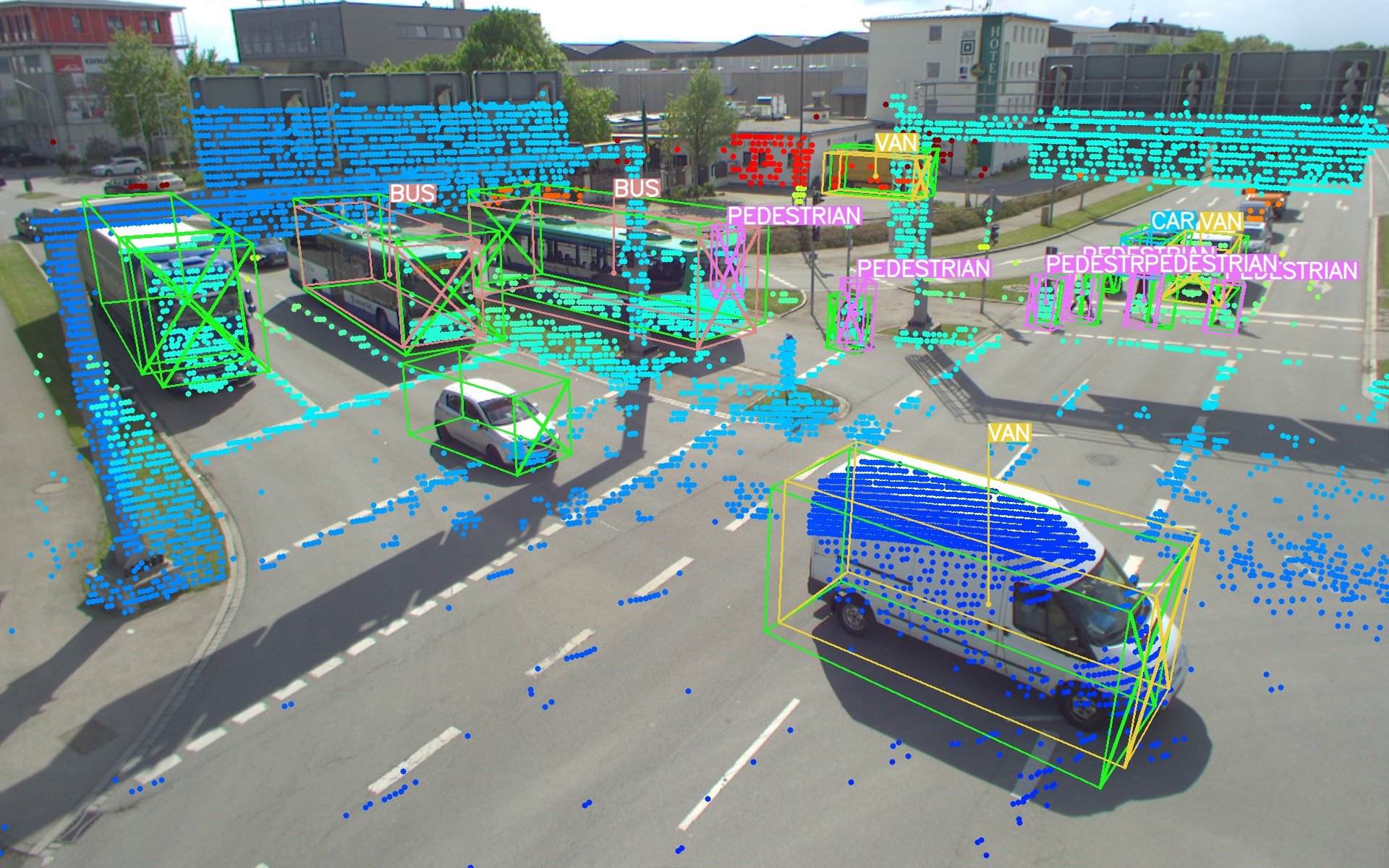}
\endminipage
\minipage{0.33\textwidth}%
  \includegraphics[width=\linewidth]{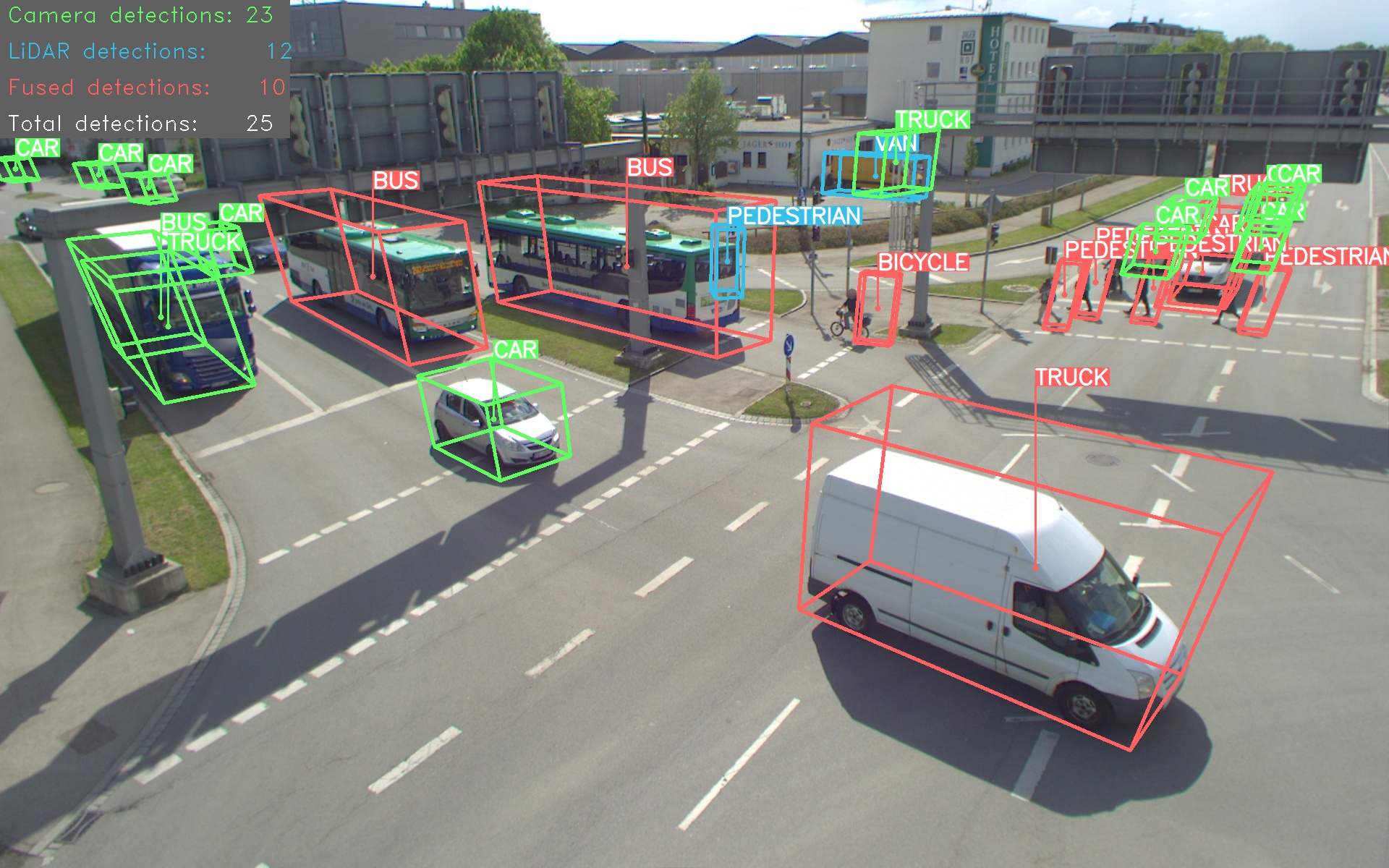}
\endminipage
\hfill
\minipage{0.33\textwidth}
  \includegraphics[width=\linewidth]{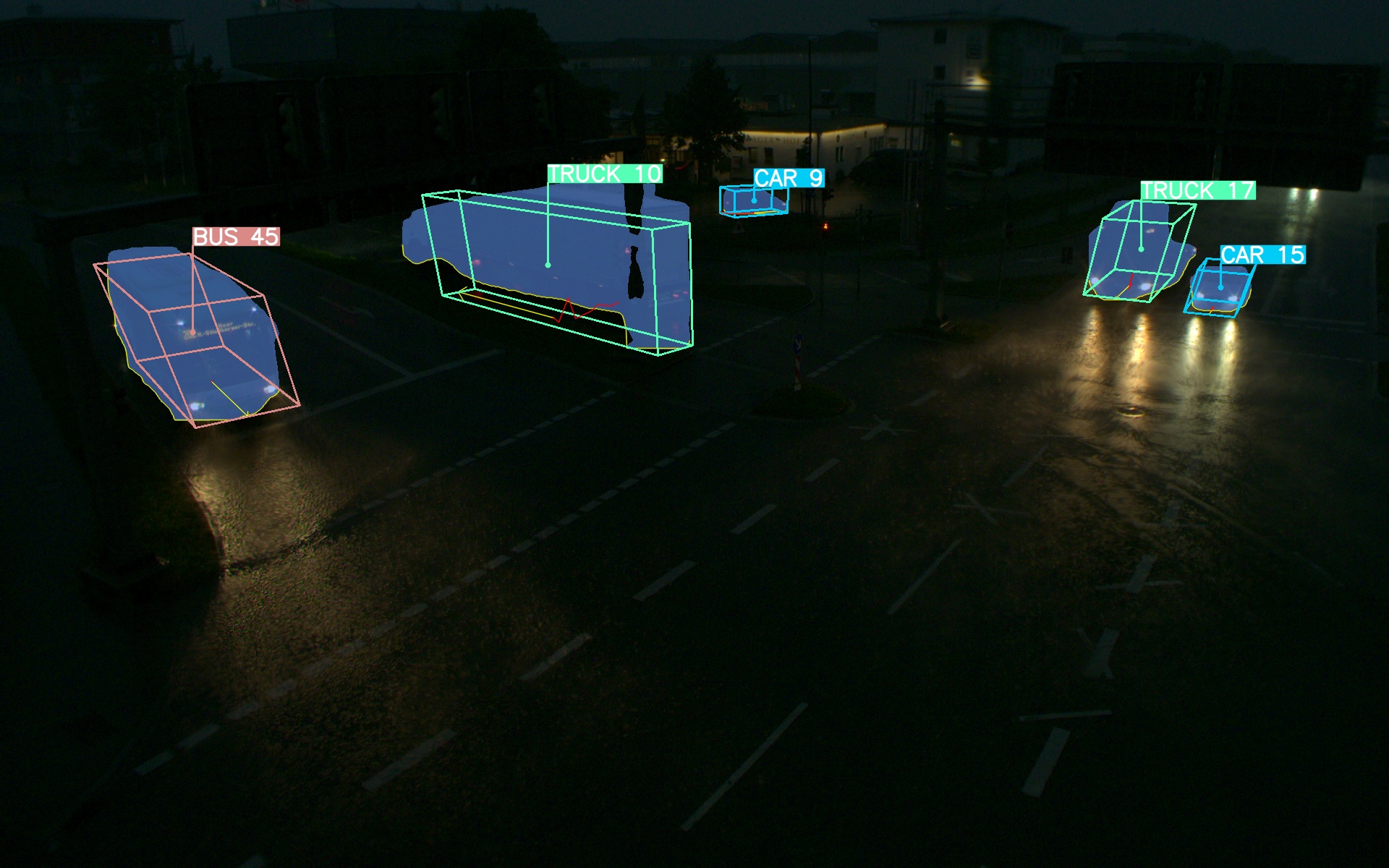}
  \caption*{(a) MonoDet3D}
\endminipage
\minipage{0.33\textwidth}
  \includegraphics[width=\linewidth]{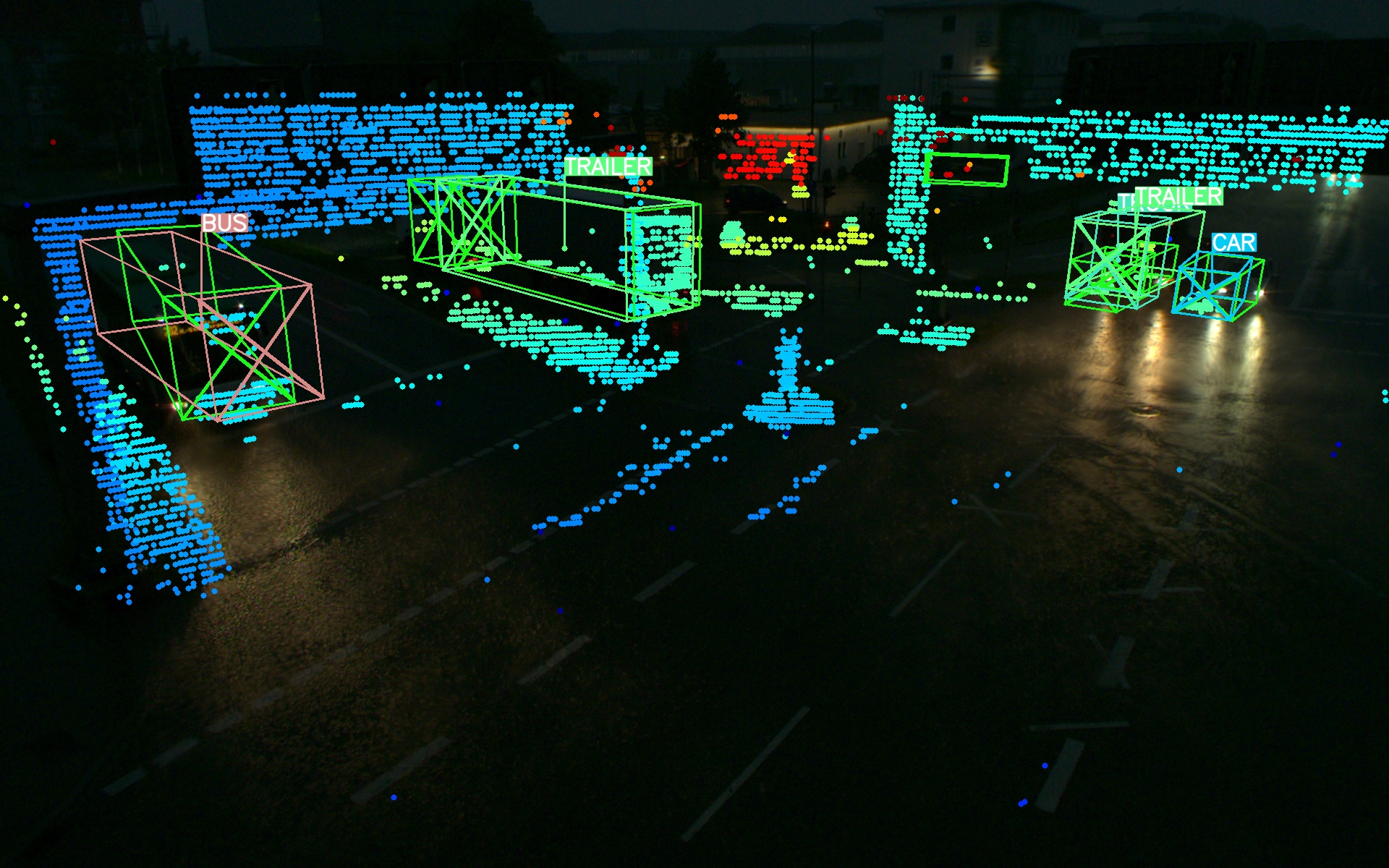}
  \caption*{(b) PointPillars}
\endminipage
\minipage{0.33\textwidth}%
  \includegraphics[width=\linewidth]{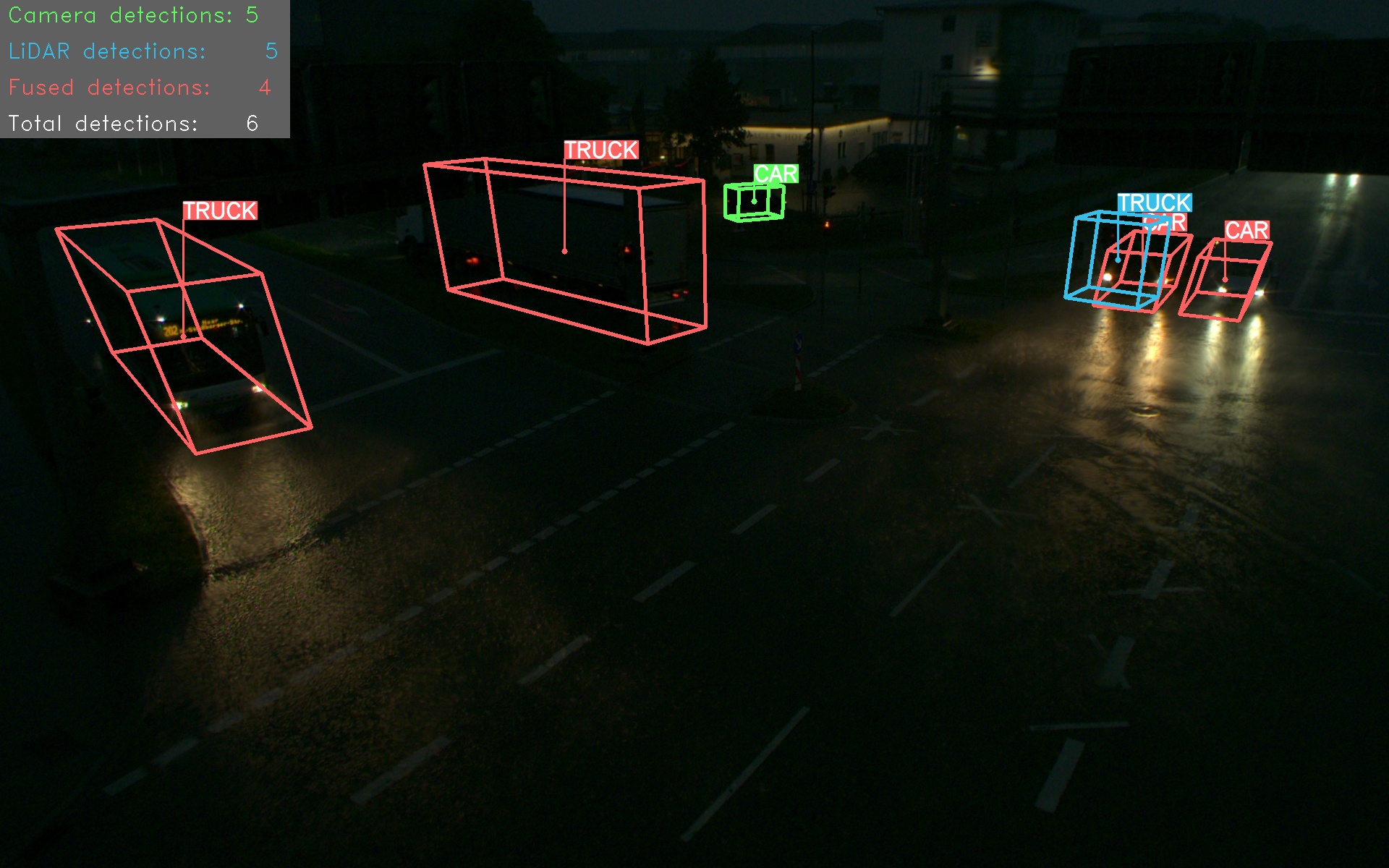}
  \caption*{(c) InfraDet3D}
\endminipage
\caption{Qualitative results on the test sequence of the three baselines: MonoDet3D (camera-only), PointPillars (LiDAR-only) and InfraDet3D (fusion) during the day (top row) and the night time (bottom row). Detections for \textit{MonoDet3D} and \textit{PointPillars} are colored by their class color. The fusion results of the \textit{InfraDet3D} model are shown in red (fused detections), green (unmatched camera detections), and blue (unmatched LiDAR detections).}
\label{fig:qualitative_results} 
\end{figure*}

In our study, we conducted a comparative analysis of monocular camera and LiDAR 3D object detection with early and late fusion. In our first evaluation experiment, we used our \textit{MonoDet3D} \cite{zimmer2023infra} 3D object detector that takes camera images as input. It transforms the 2D instance masks into 3D bottom contours by using extrinsic calibration data. Our augmented \textit{L-Shape-Fitting} algorithm extracts the dimensions and calculates the rotation for each object. In our second experiment, we used \textit{PointPillars} \cite{lang2019pointpillars} and trained the model from scratch on all classes of our camera field of views \textit{Camera\_south1}, \textit{Camera\_south2}, and \textit{full}. In the last experiment, we evaluate our multi-modal \textit{InfraDet3D} \cite{zimmer2023infra} detector, which incorporates a late fusion approach, leveraging the \textit{Hungarian} algorithm to establish correspondences between detections obtained from the \textit{MonoDet3D} and \textit{PointPillars} baselines. For all these experiments, we provide post-processing scripts in our A9-Devkit for early data fusion, and cropping the point cloud labels to fit the mentioned field of view. 

We evaluated each detector on three difficulty levels \textit{Easy}, \textit{Moderate}, and \textit{Hard}, see \Cref{tbl:evaluation}. The \textit{Hard} category contains objects with a distance over $50~\text{m}$, objects that are mostly occluded, or objects that have less than 20 points within the 3D box. Partially occluded objects with a distance of $40~\text{to}~50~\text{m}$, and 20 to 50 points are part of the \textit{Moderate} category. Lastly, the \textit{Easy} category contains objects that are not occluded, less than $40~\text{m}$ away, and contain more than 50 points. As a quantitative metric, we used the mean Average Precision (mAP) to evaluate the performance. The \textit{overall} mAP is the average of \textit{Easy}, \textit{Moderate}, and \textit{Hard}.

The advantage of using a monocular setup is a better detection of small objects such as pedestrians. On the other side, a LiDAR detector can detect objects during nighttime. The combination of LiDAR and the camera through late fusion techniques can significantly enhance the overall performance. In this work, we were able to confirm this assumption in our evaluation. We achieved the best detection results with the LiDAR\_North modality and the \textit{InfraDet3D} model in the \textit{Easy} difficulty level. Interestingly, the early fusion approach with \textit{PointPillars} consistently achieves the best performance in all subsets at \textit{Moderate} difficulty level. The better performance of \textit{PointPillars} and \textit{InfraDet3D} over the \textit{MonoDet3D} shows the strengths of the LiDAR sensor in comparison to a camera. Mostly, the late fusion of LiDAR and the camera provided better overall results than a single LiDAR detector. Moreover, the combination of early fusion between LiDAR sensors with camera sensors via late fusion, which combines the advantages of both sensor modalities, gives consistently robust results. A visual representation of the qualitative results is provided in Figure \ref{fig:qualitative_results}.

	\section{CONCLUSIONS}
In this work we extended the A9 Dataset with labeled data of an intersection. We provided 3D box labels from elevated road side sensors. Two synchronized cameras and LiDARs were used to record challenging traffic scenarios. Our data was labeled by experienced experts. As all sensors were calibrated to each other, we can use the 3D bounding box point cloud labels to perform Monocular 3D object detection. In total, our dataset contains $4.8 \text{k}$ RGB images and $4.8 \text{k}$ LiDAR point cloud frames with $57.4 \text{k}$ high-quality labeled 3D boxes, partitioned into ten object classes of traffic participants. We offered a comprehensive statistics of the labels including their occlusion levels, the number of points grouped by class category and distance, and an extensive analysis of the labeled tracks. In our evaluation experiments, we provided three baselines for the perception task of 3D object detection: A camera, a LiDAR and a multi-modal camera-LiDAR combination. With these experiments, we were able to show the potential of our dataset for your 3D perception tasks. 

For future work, we plan to create and publish more ground truth labels based on the presented camera images which can support more evaluation methods for our data fusion algorithm. Furthermore, the publication of further labeled sensor data with specific traffic scenarios, e.g. accidents, as well as the usage of other sensor modalities is also on our agenda.

	\section*{ACKNOWLEDGMENT}
	This research was supported by the Federal Ministry of Education and Research in Germany within the project $\text{\textit{AUTOtech.agil}}$, Grant Number: 01IS22088U. We thank Venkatnarayanan Lakshminarasimhan and Leah Strand for the collective work on the A9 Intersection (A9-I) Dataset.

	\bibliographystyle{IEEEtran}
	\bibliography{references.bib}

\end{document}